\documentclass[letterpaper, 10 pt, conference]{ieeeconf}  

\IEEEoverridecommandlockouts                              

\usepackage{amsmath,amssymb}
\usepackage[ruled,vlined]{algorithm2e}
\usepackage{graphicx}
\usepackage[english]{babel}

\DeclareMathOperator*{\argmax}{arg\,max}

\usepackage{subcaption}

\usepackage{biblatex}
\usepackage{lipsum}  

\title{\LARGE \bf
Mesh Based Analysis of Low Fractal Dimension Reinforcement Learning Policies
}

\author{Sean Gillen$^{1}$ and Katie Byl$^{2}$
\thanks{Sean Gillen and Katie Byl are with the Electrical and Computer Engineering Department at the University of California, Santa Barbara CA 93106
{\tt\small sgillen@ucsb.edu},
{\tt\small katiebyl@ucsb.edu}. \newline
Code hosted at: {\tt\small github.com/sgillen/fractal\_mesh}
}}

\begin{document}

\maketitle
\thispagestyle{empty}
\pagestyle{empty}

\begin{abstract}

In previous work, using a process we call meshing, the reachable state spaces for various continuous and hybrid systems were approximated as a discrete set of states which can then be synthesized into a Markov chain. One of the applications for this approach has been to analyze locomotion policies obtained by reinforcement learning, in a step towards making empirical guarantees about the stability properties of the resulting system. In a separate line of research, we introduced a modified reward function for on-policy reinforcement learning algorithms that utilizes a "fractal dimension" of rollout trajectories. This reward was shown to encourage policies that induce individual trajectories which can be more compactly represented as a discrete mesh. In this work, we combine these two threads of research by building meshes of the reachable state space of a system subject to disturbances and controlled by policies obtained with the modified reward. Our analysis shows that the modified policies do produce much smaller reachable meshes. This shows that agents trained with the fractal dimension reward transfer their desirable quality of having a more compact state space to a setting with external disturbances. The results also suggest that the previous work using mesh based tools to analyze RL policies may be extended to higher dimensional systems or to higher resolution meshes than would have otherwise been possible.

\end{abstract}

\section{INTRODUCTION}




Legged robots have clear potential to play an important role in our society in the near future. Examples include contact-free delivery during a pandemic, emergency work after an environmental disaster, or as a logistical tool for the military. Legged robots simply expand the reach of robotics when compared to wheeled systems. However, compared to wheeled systems, designing control policies for legged systems is a much more complex task, especially in the presence of disturbances, noise, and unstructured environments.

The increasing availability of massive quantities of computation has led to a resurgence of reinforcement learning (RL) in recent years. RL provides a promising approach for complex, under-actuated, hybrid control problems, such as those involved in designing control for legged locomotion. Recent examples in the context of robotics include controlling a 47 DOF humanoid to navigate a variety of obstacles \cite{heess_emergence_2017}, dexterously manipulating objects with a 24 DOF robotic hand \cite{openai_learning_2018}, and allowing a physical quadruped robot to run \cite{hwangbo_learning_2019}, and recover from falls \cite{lee_robust_2019}.

Despite the obvious promise of RL approaches, several problems need to be resolved before these systems are ready for real world applications. One of the biggest problems is that the resulting policy is typically a complete black box, there are no good ways to make theoretical, or even empirical guarantees about the resulting policies. Prior work has used so called mesh based techniques to this end \cite{Taleledeep}. Broadly, these techniques take a continuous system and approximate it with a discrete set of states. This allows us to model the system as a Markov chain, these systems are arguably easier to reason about, and it opens up a new box of tools we can bring to bear on the problem. For example we can use value iteration to switch between several controllers to improve the robustness \cite{Talelepush} or agility \cite{Byl2017} of the system. We can also perform eigen-analysis on the Markov chain's transition matrix, which provides us insights on the stability of the system \cite{Byl2009}. These techniques could both be used for policy refinement, and/or for verification and analysis of existing policies. 

However, these methods suffer from the "curse of dimensionality", because the number of possible states in our mesh grows exponentially with the degrees of freedom in our system. That is, if we make a change to the volume of continuous space represented by each discrete state, the number of states in our new mesh will grow exponentially with respect to the change in discrete state size. However for virtually all plausible walking controllers, the reachable state space is a small fraction of the total state space. Although the scaling for meshes of the reachable state space also scale exponentially as we increase the mesh resolution, the rate of scaling is typically much smaller. The scaling factor for the reachable mesh can be seen as a \textbf{fractal dimension}, which is elaborated upon in section \ref{sec:fracdim}.

In previous work \cite{Gillen2020ExplicitFractal}, we introduced a modified reward function for on-policy reinforcement learning algorithms. The reward explicitly encourages policies which induce trajectories which have a smaller fractal dimension. It's worth noting that although each individual trajectory was encouraged to have a smaller fractal dimension, this does not obviously extend to properties of the entire reachable state space for the system, which is what was used for the previous mesh based analysis of RL policies. 

In this work we take the next step and construct reachable state space meshes of agents trained with and without our modified reward. Our primary contribution is showing that these modified policies result in significantly smaller reachable meshes for a given box size, and in smaller fractal dimensions for the reachable state space. We then use the modified policies to construct a much finer mesh than would be possible otherwise. We use this mesh to compute a quantity called the mean first passage time (MFPT), and validate the obtained MFPT with Monte Carlo trials. Finally we use our mesh to produce interesting visualizations of failure states, which motivates future work.

\section{BACKGROUND}

In this section we introduce fractional dimensions, meshing, reinforcement learning, and our test environment. The environment is a hopping robot, coupled with a specific reward function. Reinforcement learning is used to train a control policy for this system which attempts to maximize the given reward function. In previous work we used a fractional dimension to modify the reward function, this modified reward results in policies that can be meshed significantly more efficiently.

\subsection{Meshing and Fractal Dimensions}
\label{sec:fracdim}

\begin{figure}[h!]
  \centering
  \begin{subfigure}[b]{0.45\linewidth}
    \includegraphics[width=\linewidth]{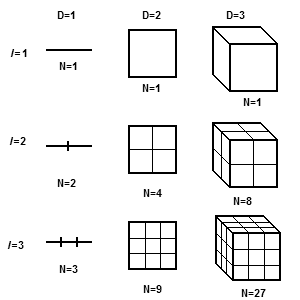}
    \caption{Scaling in different dimensions}
  \end{subfigure}
  \begin{subfigure}[b]{0.45\linewidth}
    \includegraphics[width=\linewidth]{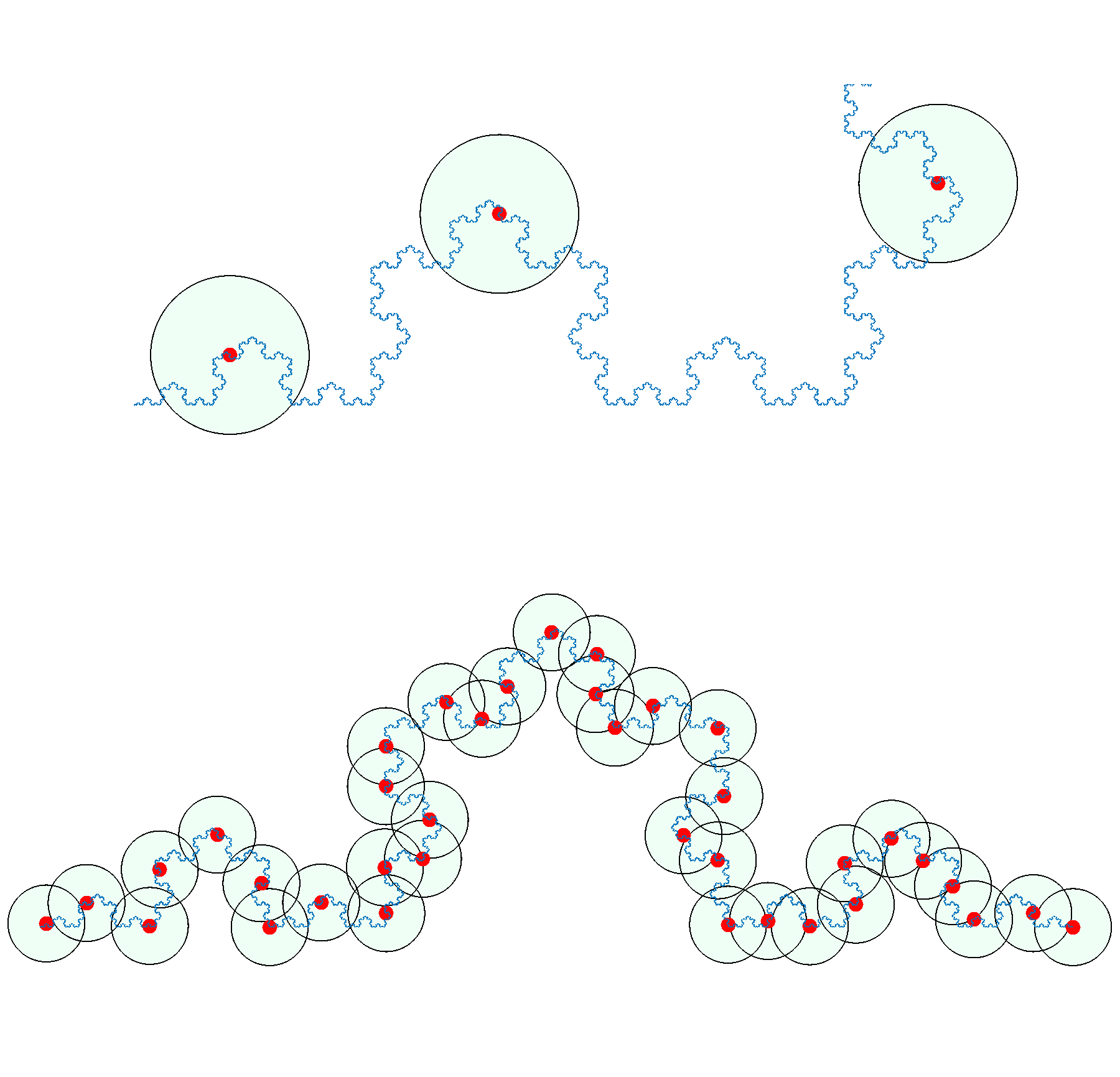}
    \caption{A non uniform mesh}
  \end{subfigure}
  \caption{Image credit: \cite{BrendanRyan/Publicdomain2020}, \cite{Talelepush}.}
  \label{fig:fracdim}
\end{figure}


Let's say we have a continuous set S that we want to approximate by selecting a discrete set M  composed of regions in S. We will call this set M a mesh of our space. Figure \ref{fig:fracdim}(a) shows some examples of this: a line is broken into segments, a square into grid spaces, and so on. The question is: as we increase the resolution of these regions, how many more regions N do we need? Again, Figure \ref{fig:fracdim}(a) shows us some very simple examples. For a D dimensional system, if we go from regions of size d to d/k, then we would expect the number of mesh points to scale as $N \propto k^{D}$. But not all systems will scale like this, as Figure \ref{fig:fracdim}(b)
illustrates. Figure \ref{fig:fracdim}(b) is an example of a curve embedded in a two dimensional space. The question of how many mesh points are required must be answered empirically. Going backwards, we can use this relationship to assign a notion of "dimension" to the curve. 

\begin{equation}
    D_{f} = -\lim_{k \rightarrow 0}\frac{\log N(k)}{\log k}. 
    \label{eq:frac_dim}
\end{equation}

This quantity is known as the Minkowski–Bouligand dimension, also called the box counting dimension. This dimension need not be an integer, hence the name "fractional" or "fractal" dimension. This is one of many measures of fractional dimensionality that emerged from the study of fractal geometry. Although these measures were invented to study fractals, they can still be usefully applied to non-fractal sets. For non fractal sets, we use the slope of the log-log relation of mesh sizes to d to compute the dimension, rather than taking a limit.



\subsection{Box Meshing}

In this work, we identify any state \textit{s} with a key obtained by: 

\begin{gather}
\nonumber s_{k} = \frac{s - \mu_{s}}{\sigma_{s}} \\ 
\text{key} = \text{round}(\frac{s_{k}}{d_{thr}})d_{thr}.   
\label{eq:key}
\end{gather}

where $\mu_{s}$ and $\sigma_{s}$ are the mean and standard deviation of all the states seen by the policy of interest during training. The round function here performs an element-wise rounding to the nearest integer. We can then use these keys to store mesh points in a hash table. Using this data structure, we can store the mesh compactly, only keeping the points we come across, and lookups are done in constant time. The parameter $d_{thr}$ is called the \textbf{box size}. Geometrically we can think of this operation as dividing the state space into a uniform grid of hypercubes, each with a side length of $d_{thr}$  

\subsection{Reinforcement Learning}

The goal of reinforcement learning is to train an agent acting in an environment to maximize some reward function. At every timestep $t \in \mathbb{Z}$, the agent receives the current state $s_{t} \in R^{n}$, uses that to compute an action $a_{t} \in \mathbb{R}^{b}$, and receives the next state $s_{t+1}$, which is used to calculate a reward $r : \mathbb{R}^{n} \times \mathbb{R}^{m} \times \mathbb{R}^{n} \rightarrow \mathbb{R}$. The objective is to find a policy  $\pi_{\theta}: \mathbb{R}^{n} \rightarrow \mathbb{R}^{m}$ that satisfies: 

\begin{equation} \argmax_{\theta} \mathop{\mathbb{E}}_{\eta}\left[ \sum_{t=0}^{T}r(s_{t}, a_{t}, s_{t+1}) \right]. \end{equation}

Where $\theta \in \mathbb{R}^{d}$ is a set that parameterizes the policy, and $\eta$ is a parameter representing the randomness in the environment. This includes the random initial conditions for episodes.

In \cite{Gillen2020ExplicitFractal}, we introduced a modified reward function:

\begin{equation} 
\argmax_{\theta} \mathop{\mathbb{E}}_{\eta}\left[ \frac{1}{D_{m}(s)}\sum_{t=0}^{T}r(s_{t}, a_{t}, s_{t+1}) \right]
\label{eq:frac_reward}
\end{equation}

where $D_{m}$ is the "lower mesh dimension" explained in detail in \cite{Gillen2020ExplicitFractal}, which is an estimate of equation \ref{eq:frac_dim} dimension for our non fractal set. 

\subsection{Environment}

\begin{figure}
\centering
\includegraphics[width=.5\linewidth]{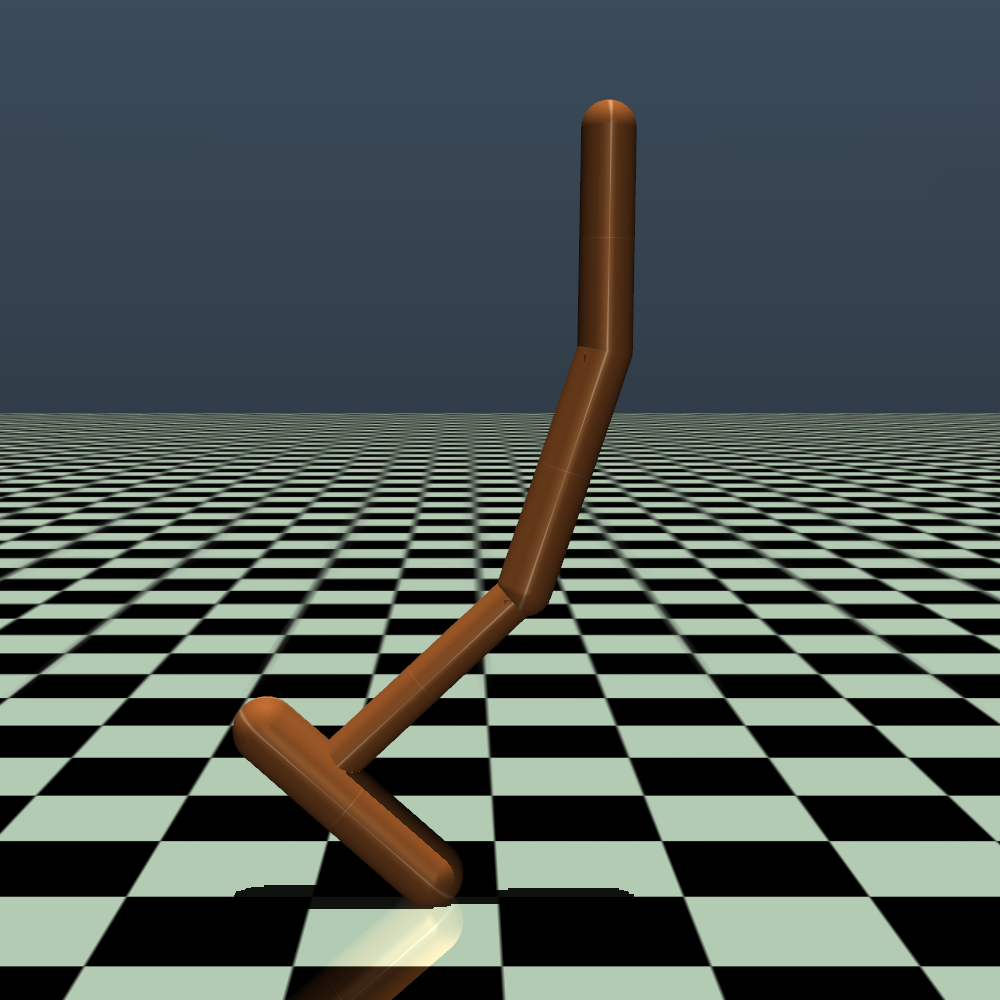}
\caption{A render of the hopper system studied in this work.}
\label{fig:hopper}
\end{figure}

Our model system is openAI gym's Hopper-v2 environment introduced in \cite{1606.01540}. This environment is part of a popular and standardized set of benchmarking tasks for reinforcement learning algorithms. The system is a 4 link, 6 DOF hopper constrained to travel in the XZ plane, seen in Figure \ref{fig:hopper}. The observation space for the agent has 11 states, the position in the direction of motion is held out, since we seek a policy that is invariant to forward progress. The actions in this case are commanded joint torques. The reward function for this environment is simply forward velocity minus a small penalty to actions. Successful controllers in this environment must execute a dynamic hopping motion to move robot along the x axis as quickly as possible. This is clearly a toy problem, but it captures many of the challenges of legged locomotion. The system is highly non-linear, under-actuated, and must interact with friction and ground contacts to maximize it's reward.


\section{MESHING}

In \cite{Gillen2020ExplicitFractal} we used meshes of individual trajectories to calculate fractional dimensions. However the previous work that has used meshing for analysis of RL policies \cite{Taleledeep} instead examines meshes of the reachable state space of a system. In particular \cite{Taleledeep} examines the reachable state space with a fixed control policy subject to a given set of disturbances. We will now outline the process for this style of meshing.  

We are interested in the set of states that our system can transition to with a fixed policy and a given set of push disturbances. We first introduce a failure state to the mesh. The failure state is assumed to be absorbing, once the robot falls it is assumed to stay that way. For our hopper, any state where the COM falls below .7m is considered to have failed, which works well in practice. This is also the failure condition of the environment during training, and therefore the agent is never trained in regions of the state space that satisfy the failure condition. 

In addition to the reachable set of states, we want to construct a state to state transition map. That is, for a given initial state, we wish to know which state we transition to for every disturbance in our disturbance set. It's worth emphasizing that this map is completely deterministic.

To make this concrete, recall that we manifest our mesh as a hash table. The key for any given state is obtained by \ref{eq:key}. When we insert a new key into our hash table, the value we place is a pair with a unique state ID (which is simply the number of keys in the table at the time of insertion), and an initially empty list of all mesh states which are reachable after one step from the key state. This data structure will provide both the reachable set, and the transition mapping.


For the hopper in particular, the system transitions from its initial standing position to a stable long term hopping gait. After letting the system enter its gait, we start detecting states on the  Poincar\'e section by selecting the state corresponding to the peak of the base link's height in every ballistic phase. These states are then collected as the initial states to seed the mesh with. Throughout this paper, we seed the mesh with trajectories from 10 initial conditions. 


For each snapshot, we initialize the system in the snap-shotted state. For each disturbance in our fixed disturbance set, we simulate the system forward subject to that disturbance. If the system does not fail, then the next Poincar\'e snapshot is captured, this state is then checked for membership in our mesh. If the new state is already in our mesh, then we simply append the new state to the list of states that the initial state can transition to. If the new state is not already in our mesh, then we expand our mesh to include the new state, and append this new state to the transition list of the initial state. If the system does fail, then we simply append the failure state to the transition list of the initial state, and no new state is added to the mesh. 

For every new state added to the mesh, we repeat this process until every state has been explored. Algorithm \ref{algo:createMesh} details this process in pseudo code.

\begin{algorithm}
\SetAlgoLined
\textbf{Input:} Initial states $S_{i}$, Disturbance set $D$\\
\textbf{Output:} Mesh M. \\
Q $\leftarrow S_{i}$ (excluding the failure state) \\
\While{Q not empty}{
    pop q from Q \\ 
    \For{d $\in$ D}{
        Initialize system in state q \\
        Run system for one step subject to disturbance d \\
        Obtain final state x \\
        \If{x $\notin$ M}{
            M[x] = List() \\ 
            Push x onto Q
        }
        Append x to M[q]
    }
}
\textbf{Return:} M \\
\caption{createMesh}
\label{algo:createMesh}
\end{algorithm}

\subsection{Stochastic Transition Matrix}

The stochastic transition matrix $\mathbf{T}$ is defined as follows:

\begin{equation}
 \mathbf{T}_{ij} = \text{Pr}(id[n+1] = j \ | \ id[n] = i)
\end{equation}

where $id[n]$ is the index in our mesh data structure of the state at step n. For some intuition, consider the transition matrix as the adjacency matrix for a graph. There is one row/column for every state in our mesh, for a given row i, each entry j is the probability of transitioning from state i to state j. Every row will sum to one, but the sum for each column has no such constraint. After constructing a mesh using algorithm \ref{algo:createMesh}, it is straightforward to create the stochastic transition matrix by iterating through every transition list in our mesh. 

\subsection{Mean First Passage Time}

We wish to use our mesh based methods to quantify the stability of our system. To do this we estimate the average number of steps the agent will take before falling, subject to a given distribution of disturbances. To do this we will use the so called Mean First Passage Time (MFPT) which in this case will describe expected number of footsteps, rather than the number of timesteps to failure. First recall that our assumption is that our failure state is an absorbing state in our Markov chain approximation, and this implies that the largest eigenvalue of \textbf{T} will always be $\lambda_{1} = 1$. In \cite{Byl2009} Byl showed that when the second largest eigenvalue $\lambda_{2}$ is close to unity, the MFPT is approximately equal to:

\begin{equation}
    MFPT \approx \frac{1}{(1-\lambda_{2}).}
    \label{eq:mfpt}
\end{equation}






\section{TRAINING}

In \cite{Mania2018} Mania et al introduce Augmented Random Search (ARS) which proved to be efficient and effective on the locomotion tasks. Rather than a neural network, ARS used static linear policies, and compared to most modern reinforcement learning, the algorithm is very straightforward. The algorithm is known to have high variance; not all seeds obtain high rewards, but to our knowledge their work in many ways represents the state of the art on the Mujoco benchmarks. Mania et al introduce several small modifications of the algorithm in their paper, our implementation corresponds to the version they call ARS-V2t, hyper parameters are provided in the appendix.

The training process is done in episodes, each episode corresponds to 1000 policy evaluations played out in the simulator. At the start of each episode, the system is initialized in a nominal initial condition offset by a small amount of noise added to each state. During each episode we fix a static policy to let the the system evolve under, we collect the observed state, the resulting action, and the resulting reward at each timestep. This information is then used to update the policy for the next episode.

We compare four different sets of agents trained in different conditions, for each training condition we use training runs across 10 different random seeds. As mentioned ARS is a very high variance algorithm, so a common practice is to run many seeds in parallel and choose the highest performing one. The standard environment has two sources of randomness which are set by the random seed. The first is a small amount of noise added to a the nominal initial condition at the beginning of each episode. The second is noise added to the policy parameters as part of the normal ARS training procedure. Using ARS in the unmodified Hopper-v2 environment will be called the \textbf{standard} training procedure. In addition to this, we have a second set of agents which are initialized with the standard training, and then trained for another 250 epochs with the fractal reward function used in equation \ref{eq:frac_reward}, these are called the \textbf{fractal agents}. Using the standard training agents as the initial policies for the fractal reward was also used in \cite{Gillen2020ExplicitFractal}, please see that manuscript for more details. 

In addition to standard training, we repeat this standard / fractal setup but with the addition of a small amount of zero mean Gaussian noise added to both the states and actions at training time. For brevity we will call these the \textbf{Standard noise} and \textbf{Fractal noise} scenarios. Hyper parameters for ARS and noise values are reported in the appendix. 
\section{Results}

\subsection{Mesh Sizes Across All Seeds}
First we wish to compare the reachable state space mesh sizes obtained for these four different training regiments. For this we assume a disturbance profile consisting of 25 pushes equally spaced between -15 and 15 Newtons, applied for 0.01 seconds along the x axis at the apex of each jump. The goal for this particular exercise is to get an idea of the relative mesh sizes among the different agents across box sizes. Table \ref{table:mesh_sizes} shows these results. We can see that across all box sizes, adding noise at run time decreases the mesh sizes slightly, and that adding the fractal reward training decreases the mesh size even further. The combination of adding noise and the fractal reward seems to perform best at reducing the mesh size.

\begin{table}[!htb]
\centering
\renewcommand{\arraystretch}{1.5}
\begin{tabular}{|l|l|l|l|l|}
\hline 
Training         & $d_{thr} = .4$ & $d_{thr} = .3$ & $d_{thr} = .2$ & $d_{thr} = .1$ \\ \hline \hline
Standard         &    64.9       &     129.0      &   289.2        &  2975.2         \\  \hline
Standard Noise   &    40.7       &     73.3       &   231.6        &  2133.3         \\  \hline
Fractal          &    26.0       &     41.8       &   67.7         &  684.4          \\ \hline
Fractal Noise    &    15.1       &     24.6       &   45.1         &  297.2          \\ 
\hline
\end{tabular}

\caption{Mesh sizes across all seeds for a disturbance profile of 25 pushes. All values are the average mesh size across 10 agents trained with different seeds.}
\label{table:mesh_sizes}
\end{table}




\subsection{Larger Meshes}

With the general trend established, we now take the best performing seed from the noisy training for further study. We chose the seed that had the smallest mesh size from both the standard noise and fractal noise agents. 

For this next experiment, we consider a richer distribution of 100 randomly generated push disturbances. These disturbances have a magnitude drawn from a uniform distribution between 5-15 Newtons. This force is applied in the xz plane with an angle drawn from a uniform distribution between 0 and 2$\pi$. The number of forces was chosen by increasing the number of forces sampled until the mesh sizes between two random sets did not change. The magnitude of the pushes was chosen arbitrarily, in principle one can use these methods for any distribution of disturbance they expect their robot to encounter during operation. 

\begin{figure}[!htb]
\centering
\includegraphics[width=.7\linewidth]{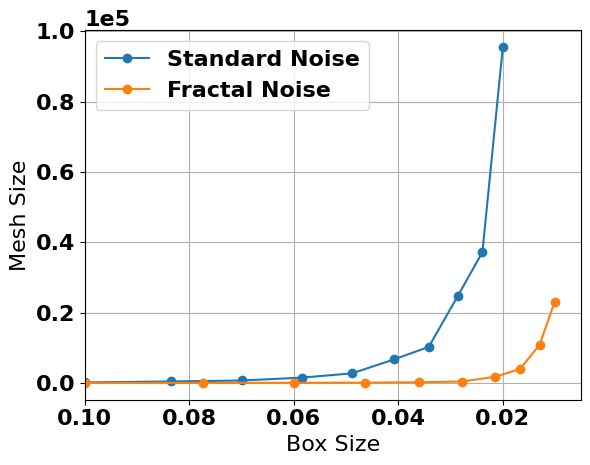}
\caption{Mesh sizes for the top performing standard noise and fractal noise agents.}
\label{fig:mesh_cmp}
\end{figure}

We then construct meshes for different box sizes. For each agent we construct 10 meshes. We vary the box size between 0.1 and 0.01 for the fractal noise agent. For the standard noise agent we instead vary the box size between 0.1 and 0.02 because the mesh sizes for the standard agent were proving to be too large at the smaller box sizes. Figure \ref{fig:mesh_cmp} shows the comparison, We can see clearly that at the very least, the exponential blowup in mesh size starts at much more accurate mesh resolutions for the fractal agent.

We are also interested in the exponential scaling factor in the mesh size as the box gets smaller, which is captured by the fractal dimension discussed in section \ref{sec:fracdim}. As mentioned before, in previous work our modified reward signal resulted in agents with a smaller fractal dimension with respect to individual trajectories. We now ask if this carries over to meshes of the reachable state space obtained by the procedure from algorithm \ref{algo:createMesh}. Table \ref{table:mesh_dims} shows the results, we can see that indeed, the fractal training does seem to reduce the mesh dimensionality for the reachable state space meshes.

\begin{table}[!h]
\renewcommand{\arraystretch}{1.5}
\begin{tabular}{|l|l|l|}
\hline 
Training              & Trajectory Mesh Dim.    &  Reachable Mesh Dim. \\ \hline \hline
Standard Noise        &    1.38                 &      3.83     \\  \hline
Fractal Noise         &    1.16                 &      3.16     \\  \hline
\end{tabular}

\caption{Mesh dimensions for the best performing seed from the standard with noise training, and the fractal with noise training, given the same disturbance profile of 100 pushes. For reference the state space for our system has 12 dimensions.}
\label{table:mesh_dims}
\end{table}

\subsection{Validating the Mean First Passage Time}

We emphasize that the reward function for the hopper environment is simply to move forward with the highest velocity possible, no attempts were made to make the system robust to disturbances. Perhaps because of this, the mean first passage time for these systems are relatively small, on the order of 100 foot steps. For this small number of steps, we can validate the mean first passage time with Monte Carlo trials. It's worth noting that the eigen estimate of the mean first passage time is much more valuable for more robust systems. This is because this estimate becomes more accurate as the system becomes more stable, and because the cost of calculating the MFPT with Monte Carlo trials grows much more expensive for more stable systems. In previous works \cite{OguzSaglam2015} it was used to quantify robustness for systems with a MFPT as high as $10^{15}$. 

To do this, we compare the mean first passage time as estimated by equation \ref{eq:mfpt} to the value computed by looking at many Monte Carlo rollouts. For the rollouts we apply a random action drawn from the same distribution described above. Instead of sampling 100 pushes though we sample a new push every time we need a new disturbance. During the rollouts we still apply the push at the apex height of the ballistic phase.

\begin{figure}[!h]
\centering
\includegraphics[width=.75\linewidth]{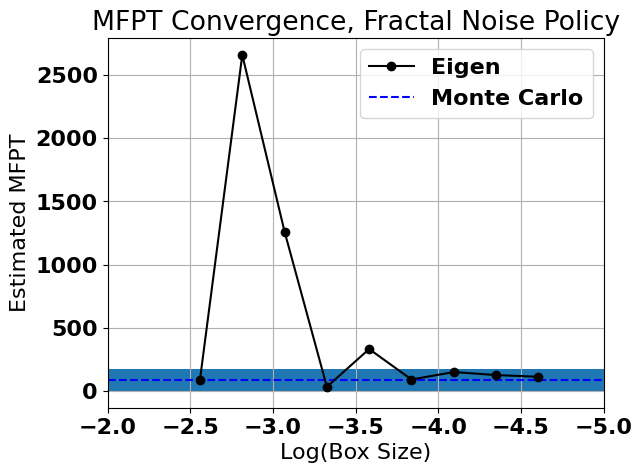}
\caption{Estimated mean first passage time computed from \ref{eq:mfpt} compared to a Monte Carlo estimate. The blue dashed line and shaded region are the mean and standard deviation of the steps to failure for 2500 Monte Carlo rollouts.}
\label{fig:mdim_monte}
\end{figure}

Figure \ref{fig:mdim_monte} shows the convergence of the MFPT as we expand the size of the mesh, and compares it to the mean steps to failure obtained with Monte Carlo trials. We can see that it does look like the MFPT is converging to the Monte Carlo result. Although at the largest mesh we tried, the eigen analysis gives an estimate of 110.2 steps to failure, while the Monte Carlo trials tell us that an average of 85 steps are taken before failure. It's worth noting that the distribution of failure times has a large variance with a standard deviation of 80 steps.

\subsection{High Resolution Mesh}

\begin{figure}[!htb]
\centering
\includegraphics[width=.7\linewidth]{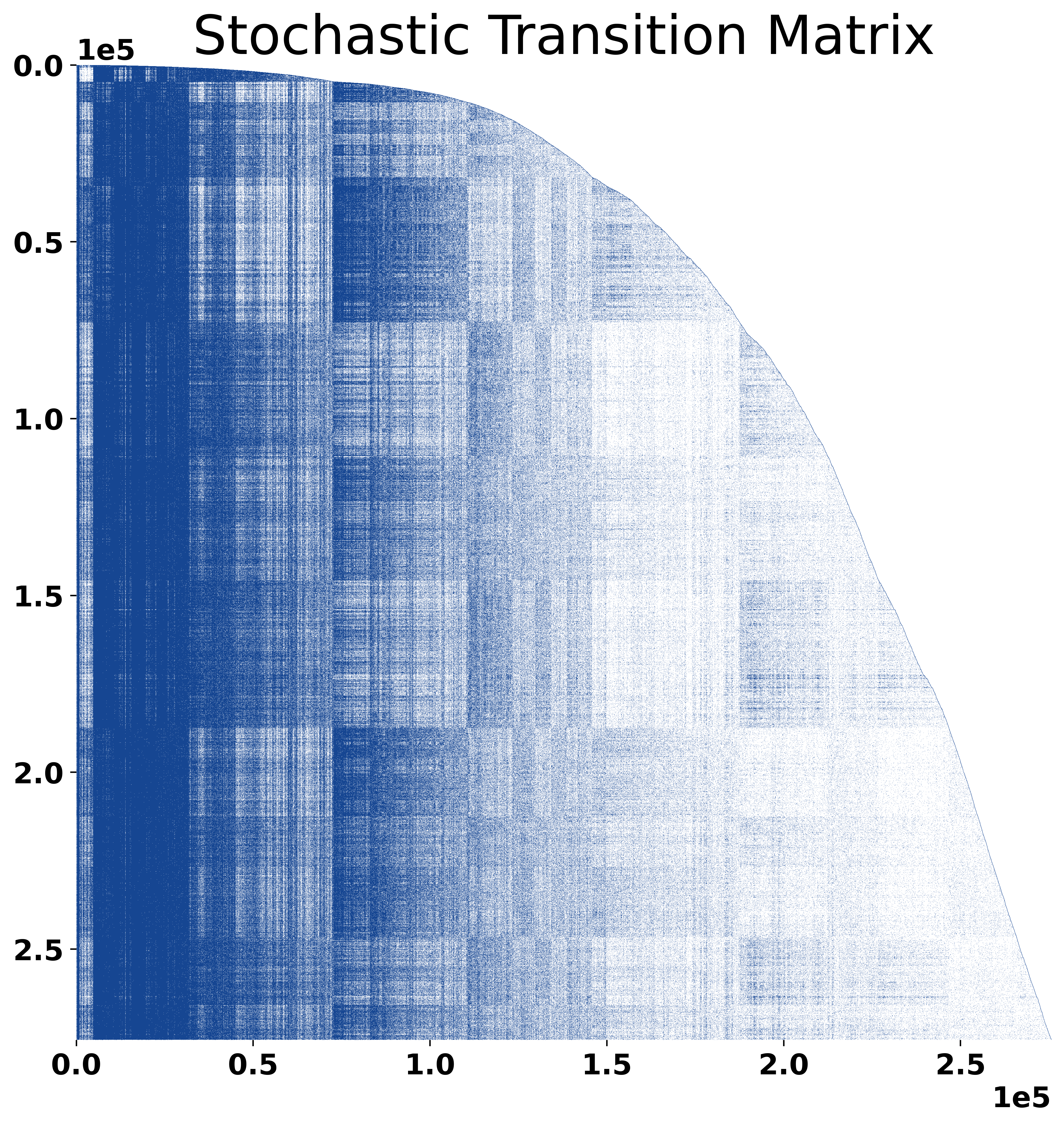}
\caption{Visualization of the stochastic transition matrix for the top performing fractal noise agent. All non zero values are shown with equal size and coloration. Recall that each entry in $T_{ij}$ tell us the probability of transitioning to state j after one step if we start in state i.}
\label{fig:mdimT}
\end{figure}

We now use the fractal agent and construct an even more accurate mesh. Figure \ref{fig:mdimT} show the sparsity pattern for the state transition matrix for the fractal noise agent with a box size of 0.005. Recall that in the process for creating the mesh, we start with a small number initial seed states. After that every new state that we add is added in order we find them to the mesh. So if we are expanding state \# 2, and there are currently 100 states in the mesh, if we transition to an unseen state, that state will be labeled \# 101. So although it may seem like it is not possible for states in the top right quadrant to visit states later in the mesh, this is really an artifact of how we construct our mesh and label our points. 


\begin{figure}[!htb]
\centering
\includegraphics[width=.7\linewidth]{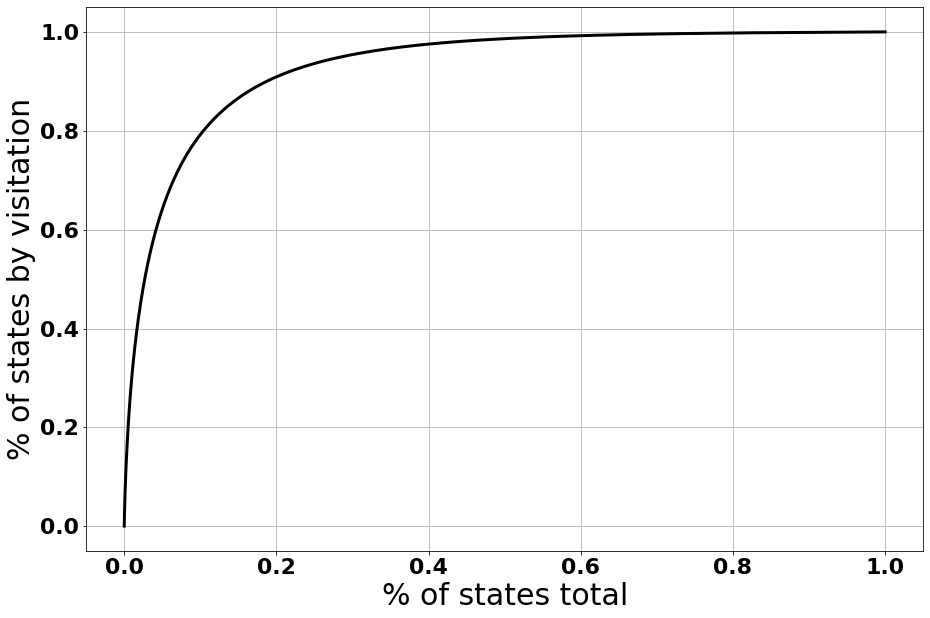}
\caption{Cumulative sum of probability mass excluding the failure state. We take the sum of each column of T, and sort it in descending order, then report the cumulative sum of probability. Each point on the curve tells us that x\% of states make up y\% of all state transitions.}
\label{fig:mass_clump}
\end{figure}

We note that there are a smaller set of states that make up most of the transitions. In fact we can see from Figure \ref{fig:mass_clump} that 20\% of the states in our mesh account for about 90\% of all transitions seen during the mesh construction.

\begin{figure}
\centering
\includegraphics[width=\linewidth]{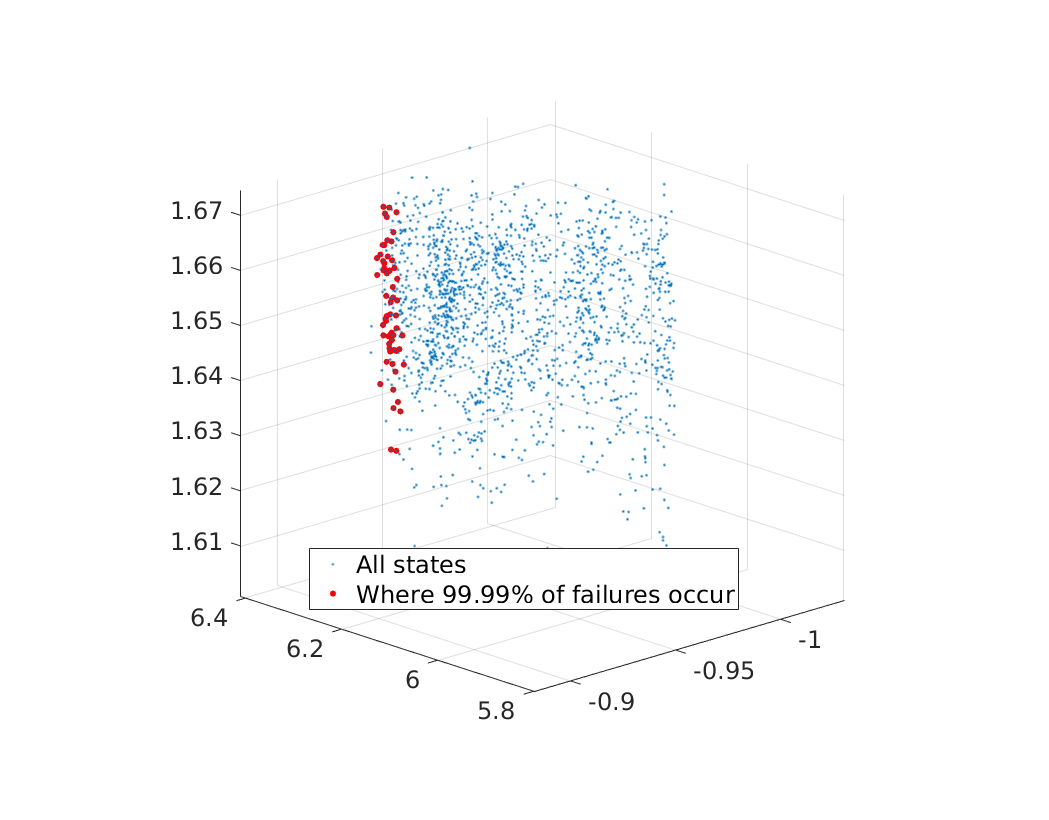}
\caption{View of the first 3 principle components of the mesh for a fractal noise policy.}
\label{fig:pca_side}
\end{figure}

One of the advantages of having a discrete set of states is that it opens up new tools and visualizations, for example we can apply Principle Component Analysis (PCA). Figure \ref{fig:pca_side} shows a projection of our mesh states on the top 3 principle components. We note that these three states account for more than 97\% of the variance, we also note that our analysis reveals that states in red are where 99\% of all failures occur. The visualization reveals that at least in PCA space, all the trouble states are clustered in one spot. A promising direction for future work is to introduce a policy refinement step that attempts to avoid these states. Additionally, if we were designing a real robot this may give us insights into design changes that could be made.

\section{CONCLUSIONS}

In this work, we apply previously developed tools that create discrete meshes for the reachable state space of a system. These tools were applied to policies obtained with a modified reinforcement learning reward function which was previously shown to encourage small mesh dimensions for individual trajectories not subject to any disturbances. We showed that these modified policies have a smaller average reachable mesh size across all random seeds for coarse meshes and a small number of disturbances. We then showed a clear difference in mesh sizes and mesh dimensions for the top performing seeds on a richer set of disturbances and finer mesh sizes. We also validated our use of the MFPT as a tool by comparing it to Monte Carlo trials. Finally, we constructed a high fidelity mesh at a resolution that would not have been feasible with standard ARS policies. In addition, we created visualizations with this mesh that revealed insights about the contracting nature of the policy, and which point to future applications of this approach. Taken together, these results show two things. First, it further validates the utility of the fractal dimension reward, which we have shown transfers it's desirable quality of having a more compact state space to a setting with external disturbances. These results are also a credit to the mesh based tools, because it shows that the fractal training can be used to extend the reach of these tools to higher dimensional systems or higher resolution meshes than would have otherwise been possible.

\section*{APPENDIX}

\subsection*{Hyper Parameters}
\textbf{ARS:} (from \cite{Mania2018}) $\alpha = 0.02$, $\sigma = 0.025$, $N=50$, $b=20$.  \\
\textbf{MeshDim:} (from \cite{Gillen2020ExplicitFractal}) f = 1.5, $d_{0}$ = 1e-2

\subsection*{Noise During Training}
Zero mean Gaussian noise with std = 0.01 added to policy actions before being passed to the environment, for reference all actions from the policy are between -1 and 1.  Zero mean Gaussian noise with std = 0.001 added to observations before being passed to the policy.  \\ 



\clearpage
\printbibliography  

\addtolength{\textheight}{-12cm}   





\end{document}